%% file: main.tex
\documentclass[11pt]{article}
\usepackage[margin=1in]{geometry}
\usepackage{amsmath,amssymb}
\usepackage{booktabs}
\usepackage{graphicx}
\usepackage{subcaption}
\graphicspath{{figures/}}
\usepackage{hyperref}
\usepackage{natbib}

\title{How Much of a 10-K Matters? Aggregation-Dependent Value of Full-Text versus Risk-Factor Sentiment}
\author{Sanggyu Sean Choi \\ Artificial Intelligence and its Application Institute \\University of Edinburgh \\ \texttt{schoi3@ed.ac.uk}}
\date{}

\begin{document}
\maketitle

\begin{abstract}

Financial sentiment extraction has largely relied on news text and supervised extraction against return labels alone, leaving 10-K filings — and volatility, the target risk disclosure is arguably best suited to informing — comparatively unexplored. We extend a supervised lexicon-learning approach to 10-K filings and their Item 1A risk-factor sections, training sentiment scores against both return and volatility labels at three levels of aggregation: sector, portfolio, and individual firm. Across 1,383 filings from 94 Nasdaq-100 technology constituents (2006–2023), we evaluate the resulting twelve sentiment metrics on classification accuracy, correlation with realised market outcomes, and qualitative lexical content. Full-filing text produces more accurate sentiment at the sector and portfolio level for both targets, but this reverses at the individual-firm level, where the narrower Item 1A section performs better — an effect we attribute to the interaction between document volume and the amount of independent training signal available at each level of aggregation. A Loughran-McDonald dictionary baseline is consistently, strongly negatively correlated with price at every level tested, underscoring the value of a supervised approach for regulatory disclosure text. These findings, and the design choices they motivate, establish the sentiment-generation methodology underlying a subsequent, larger-scale, multi-source system.
\end{abstract}

\input{sections/intro}
\input{sections/related_work}
\input{sections/data}
\input{sections/methodology}
\input{sections/results}
\input{sections/discussion}
\input{sections/conclusion}

\bibliographystyle{plainnat}
\bibliography{references}

\end{document}

%% file: sections/intro.tex
\section{Introduction}

Corporate disclosures are a primary channel through which firms communicate
information that markets subsequently price. Among these, the 10-K filing --
an annual report mandated by the U.S. Securities and Exchange Commission -- is
distinctive in combining legally required comprehensiveness with a dedicated
risk-disclosure section (Item 1A) that firms use specifically to enumerate
forward-looking uncertainties. This makes 10-K text a natural candidate for
extracting sentiment signals relevant to both the direction of future returns
and the level of future volatility. Most prior work on sentiment-based return
prediction, however, has drawn on news text rather than regulatory filings,
and has supervised sentiment extraction against return labels alone, leaving
volatility -- arguably the quantity risk-disclosure language is most directly
suited to informing -- largely unaddressed.

This paper develops and evaluates a supervised sentiment-generation approach
for 10-K filings, extending the supervised lexicon-learning framework of
\citet{ke2020predicting} -- originally developed for return prediction from
news articles -- along three dimensions. First, we apply it to regulatory
filings rather than news text, and directly compare sentiment generated from
the full filing against sentiment generated from the Item~1A risk-factor
section alone, testing whether a targeted disclosure section carries more or
less predictive sentiment content than the filing as a whole. Second, we
supervise sentiment generation against both return and volatility labels; to
our knowledge, no prior application of this approach -- including the one it
extends -- has used volatility as a supervisory target. Third, we evaluate
the resulting sentiment metrics at three levels of aggregation: sector (the
technology sector as a whole), portfolio (a ten-firm technology portfolio),
and individual firm, allowing us to characterise how sentiment behavior
changes with the level of aggregation at which it is measured.

We evaluate the resulting sentiment scores on three independent axes: their
accuracy against held-out return and volatility labels; their correlation
with realised market outcomes, benchmarked against a Loughran--McDonald
dictionary baseline \citep{LoughranMcDonald2011}; and a qualitative analysis
of the words driving each model's scores, which we use to characterise the
underlying themes each model has learned to associate with positive and
negative sentiment.

\paragraph{Contributions.}
\begin{enumerate}
  \item A supervised, lexicon-learning approach to sentiment scoring applied
  to 10-K filings and their Item~1A risk-factor sections -- the first
  application of this methodology to regulatory filing text.
  \item A full-filing versus risk-factor-section ablation, evaluated at
  sector, portfolio, and firm level, isolating whether targeted risk
  disclosure or comprehensive filing text carries more predictive sentiment.
  \item Sentiment scores supervised against both return and volatility
  labels, extending a return-only supervision framework to a second,
  arguably more naturally suited target.
  \item A three-axis evaluation -- classification accuracy, correlation
  analysis, and qualitative lexical interpretation -- of the resulting
  twelve sentiment metrics.
\end{enumerate}

%% file: sections/related_work.tex
\section{Related Work}

\paragraph{Lexicon-based sentiment analysis.}
Financial sentiment extraction has traditionally relied on dictionary-based
scoring, in which text is matched against a predefined, polarity-labeled
word list. General-purpose dictionaries such as Harvard IV-4 misclassify a
substantial share of financial terms -- \citet{LoughranMcDonald2011} show
three-fourths of Harvard-IV-4-flagged ``negative'' words in 10-K filings do
not carry negative connotation in a financial context -- motivating
finance-specific dictionaries such as their own, which we use as an
evaluation baseline (Section~\ref{sec:results}). Dictionary methods are
transparent and require no training data, but polarity is fixed in advance
and does not adapt to the specific target being predicted.

\paragraph{Supervised and machine-learning approaches.}
Machine-learning approaches range from classical bag-of-words classifiers to
deep architectures (RNNs, CNNs, transformer-based models) capable of
capturing contextual and sequential structure, at the cost of longer
inference time, larger data requirements, and reduced interpretability.
\citet{ke2020predicting} propose an intermediate approach: a supervised
lexicon-learning model in which sentiment-charged words and their associated
topic distributions are estimated from data -- rather than fixed by a
dictionary or learned by an uninterpretable model -- using realised returns
as a training signal, applied to news text. This preserves term-level
interpretability while adapting to the prediction target and text domain,
and is the methodology this paper extends to 10-K filings.

\paragraph{Textual analysis of regulatory filings.}
A separate literature examines 10-K and related filings directly, typically
using dictionary-based scoring \citep{LoughranMcDonald2011} or
transformer-based models \citep{Blomme2020} to relate filing tone
to abnormal returns. This literature has not, to our knowledge, applied
supervised lexicon-learning to filing text, nor supervised sentiment
extraction from filings against a volatility target -- the gap this paper
addresses.

%% file: sections/data.tex
\section{Data}
\label{sec:data}

\paragraph{Universe.}
We construct our sample around the Invesco QQQ Trust Series~1 (QQQ), an
exchange-traded fund tracking the Nasdaq-100 Index, which is conventionally
used as a proxy for the U.S.\ technology sector. Of the 100 constituent
firms, 94 file Form 10-K with the SEC; the remaining six are
foreign-domiciled and file Form 20-F instead, and are excluded.

\paragraph{Retrieval.}
10-K filings are retrieved from the SEC's EDGAR system, indexed by each
firm's Central Index Key (CIK) rather than ticker symbol, with requests
throttled to EDGAR's published rate limit of ten requests per second.
Filings are retrieved in HTML format, which the large majority of EDGAR
filings use natively. In total, 1{,}383 filings were collected across 94
firms, spanning January 2006 to December 2024.

\begin{figure}[h]
\centering
\includegraphics[width=0.85\textwidth]{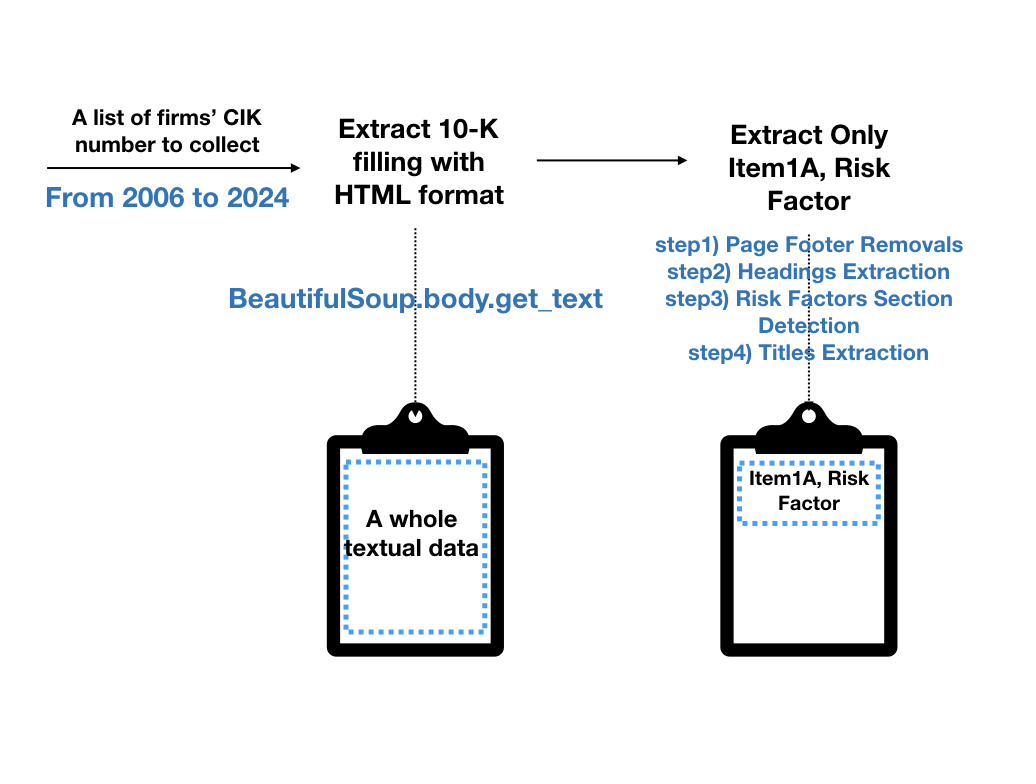}
\caption{10-K filing extraction pipeline. A list of firms' CIK numbers is
used to retrieve full 10-K filings in HTML format from EDGAR (2006--2023),
from which the Item~1A risk-factor section is isolated via a four-step
process: page footer removal, heading extraction, risk-factor section
detection, and title extraction.}
\label{fig:extraction}
\end{figure}

\paragraph{Extraction.}
Figure~\ref{fig:extraction} summarizes the extraction pipeline. For each
filing, we extract two document variants: the full filing text,
and the Item~1A risk-factor section alone. Item~1A extraction exploits four
structural regularities common to 10-K HTML: (i) non-informative page
footers (page numbers, continuation notices, disclaimers) are removed by
detecting horizontal-rule and page-break markers; (ii) heading elements are
identified via consistent HTML style attributes (bold, underline, italic
markup) and demarcated for downstream parsing; (iii) the risk-factor section
itself is located between the ``Item~1A'' heading and the next recognized
section heading (e.g., ``Item~1B,'' ``Item~2,'' or firm-specific variants
such as ``Management's Report''), using a regular-expression pattern set
tolerant of the spacing and quotation-mark variants EDGAR filings exhibit;
and (iv) section titles are distinguished from sub-headings by
capitalisation pattern once stopwords are removed. Item~1A disclosure has
been a mandatory, separately captioned section since 2005
\citep{sec2005}, which we use as the natural starting point of the
collection window.

Aggregated, post-preprocessing word counts across all 94 firms' filings
total approximately 15{,}863 (full filing) versus 9{,}030 (Item~1A alone) --
the risk-factor section, despite constituting only 10--15\% of filing
length, retains roughly 57\% of the full filing's post-preprocessing word
count, consistent with risk disclosure being disproportionately dense in
sentiment-relevant vocabulary.

\paragraph{Document-term representation.}
Extracted text is tokenised and converted into a document-term matrix
separately for the full-filing and Item~1A variants, feeding the Sentiment
Score Prediction Model described in Section~\ref{sec:methodology}.

%% file: sections/methodology.tex
\section{Methodology}
\label{sec:methodology}

\begin{figure}[h]
\centering
\includegraphics[width=\textwidth]{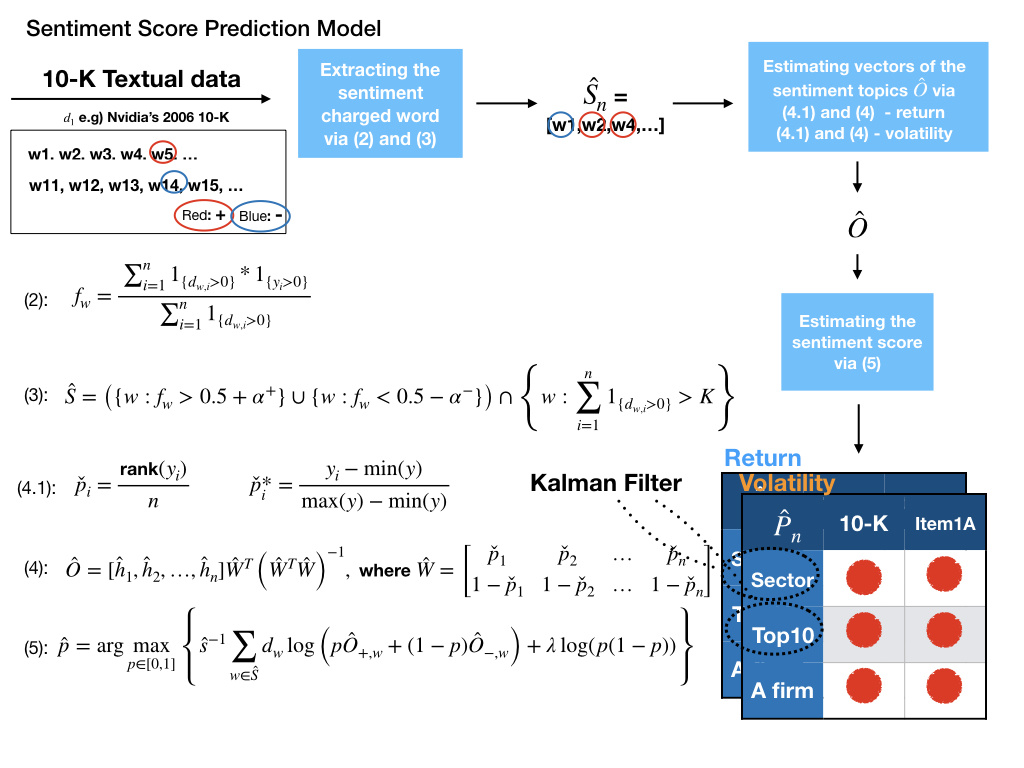}
\caption{The Sentiment Score Prediction Model. Sentiment-charged words are
extracted from 10-K textual data (Eqs.~\ref{eq:fw}--\ref{eq:shat}), used to
estimate topic-distribution vectors $\hat{O}$ under return or volatility
labeling (Section~\ref{sec:labeling}), which in turn determine the
penalized-likelihood sentiment score $\hat{p}$ (Eq.~\ref{eq:phat}) reported
across the three stakeholder levels (sector, portfolio, firm) and both
document variants (10-K, Item~1A).}
\label{fig:sentimentmodel}
\end{figure}

Let $D \in \mathbb{R}^{n \times m}_+$ be the document-term matrix over $n$
filings and vocabulary $V$ of size $m$, with row $d_i$ the word-count vector
of filing $i$. We construct $D$ separately for the full-filing and Item~1A
variants described in Section~\ref{sec:data}, yielding two independent
document-term representations per stakeholder level. Each filing $i$ is
associated with a label $y_i$ -- return or volatility, depending on the
target being trained -- measured relative to the filing's publication date,
and is assumed to have an underlying, unobserved sentiment score
$p_i \in [0,1]$. Figure~\ref{fig:sentimentmodel} summarizes the full
pipeline described in this section.

\paragraph{Model assumptions.}
The model rests on three assumptions, following \citet{ke2020predicting}.
First, the vocabulary decomposes into sentiment-charged words $S$ and
neutral words $N$ ($V = S \cup N$), with a filing's tone determined only by
$d_{[S],i}$, the word counts restricted to $S$. Second, sentiment-charged
word counts follow a mixture multinomial distribution,
\begin{equation}
d_{[S],i} \sim \text{Multinomial}\left(s_i,\ p_i O^+ + (1-p_i) O^-\right),
\label{eq:mixture}
\end{equation}
where $s_i = \sum_{w \in S} d_{w,i}$ is the total sentiment-word count in
filing $i$, and $O = [O^+, O^-] \in \mathbb{R}^{|S| \times 2}_+$ are
word-distribution vectors representing $0<=p_i<=1$ — the purest positive ($p_i=1$) and
purest negative ($p_i=0$) sentiment topics — each summing to one over $S$.
Third, the sentiment score is assumed sufficient for the relationship
between filing content and the label: $y_i \mid p_i \perp\!\!\!\perp d_i$.

\paragraph{Estimating sentiment-charged words.}
A word $w$ is scored by the fraction of its occurrences falling in filings
with a positive label,
\begin{equation}
f_w = \frac{\sum_{i=1}^n \mathbb{1}\{d_{w,i}>0\} \cdot \mathbb{1}\{y_i>0\}}
{\sum_{i=1}^n \mathbb{1}\{d_{w,i}>0\}},
\label{eq:fw}
\end{equation}
and is retained as sentiment-charged if $f_w$ is sufficiently far from $0.5$
-- governed by threshold parameters $\alpha^+, \alpha^- \in (0, 0.5]$ -- and
occurs more than $k$ times across the corpus, excluding rare words for which
$f_w$ is a noisy estimate:
\begin{equation}
\hat{S} = \Big(\{w : f_w > 0.5+\alpha^+\} \cup \{w : f_w < 0.5-\alpha^-\}\Big)
\cap \Big\{w : \textstyle\sum_{i=1}^n \mathbb{1}\{d_{w,i}>0\} > k\Big\}.
\label{eq:shat}
\end{equation}

\paragraph{Estimating topic distributions.}
Since a filing's true sentiment score is unobserved, $\hat{O}$ is estimated
using a proxy $\check{p}_i$ constructed from $y_i$
(Section~\ref{sec:labeling}). Writing $\hat{h}_i = d_{[\hat{S}],i}/\hat{s}_i$
for the relative sentiment-word frequencies and
$\hat{W} = [\check{p}_i;\ 1-\check{p}_i]_{i=1}^n$ for the proxy matrix,
\begin{equation}
\hat{O} = \hat{H}\hat{W}^\top(\hat{W}\hat{W}^\top)^{-1},
\label{eq:ohat}
\end{equation}
with negative entries clipped to zero and columns renormalized to sum to
one.

\paragraph{Scoring new filings.}
Given $\hat{S}$ and $\hat{O}$, a new filing's sentiment score is the
maximizer of a penalized log-likelihood under the mixture model of
Eq.~\eqref{eq:mixture}:
\begin{equation}
\hat{p} = \arg\max_{p \in [0,1]} \left\{ \hat{s}^{-1} \sum_{w \in \hat{S}}
d_{i,w} \log\!\big(p\,\hat{O}^+_w + (1-p)\,\hat{O}^-_w\big) +
\lambda \log\big(p(1-p)\big) \right\}.
\label{eq:phat}
\end{equation}
The penalty term shrinks $\hat{p}$ toward $0.5$, with effect largest for
filings with few sentiment-charged words -- preventing a handful of extreme
words from producing an overconfident score. We set $\lambda = 0.1$
throughout.

\subsection{Labeling}
\label{sec:labeling}

\paragraph{Return.}
The return proxy is the filing's percentile rank among all realized
returns: $\check{p}_i = \text{rank}(y_i)/n$, using open-to-close log returns
computed only from price information available up to the filing's
publication date, to avoid look-ahead bias.

\paragraph{Volatility.}
Volatility does not admit the same rank-based proxy without discarding its
asymmetric structure. We instead use a normalized proxy
$\check{p}_i^* = (y_i - \min(y))/(\max(y) - \min(y))$, and replace the $0.5$
threshold in Eqs.~\eqref{eq:fw}--\eqref{eq:shat} with a tunable pair
$(\theta, q)$ separating high- from low-volatility filings, set to the 65th
percentile and $0.65$ respectively, tuned on the training window for both
sector- and firm-level models. The volatility target is computed from the
intra-daily log price range, converted to a volatility estimate following
\citet{corsi2009simple}, and averaged over a three-day window to reduce
single-day noise.

\subsection{Kalman Filter}

Filing arrivals are irregular across firms, and multiple firms can file on
the same day, so raw daily sentiment aggregates are noisy at the sector and
portfolio level. We average per-day scores over all filings published that
day, $\bar{p}_i = |A_i|^{-1}\sum_{i \in A_i} \hat{p}_i$, and smooth the
resulting series with a local-level Kalman filter
\citep{durbin2012time}:
\begin{align}
\mu_{i+1} &= \mu_i + \eta_i, & \eta_i &\sim \mathcal{N}(0,\sigma_\eta^2), \\
\bar{p}_i &= \mu_i + \varepsilon_i, & \varepsilon_i &\sim \mathcal{N}(0,\sigma_\varepsilon^2),
\end{align}
producing a filtered series $\tilde{p}_i$ used for sector- and
portfolio-level correlation analysis (Section~\ref{sec:results}). Firm-level
scores are retained unfiltered, since firm-level analysis benefits from
preserving filing-specific variation rather than a denoised trend.

%% file: sections/results.tex
\section{Experimental Setup and Results}
\label{sec:results}

\subsection{Experimental Setup}

\begin{figure}[h]
\centering
\includegraphics[width=0.75\textwidth]{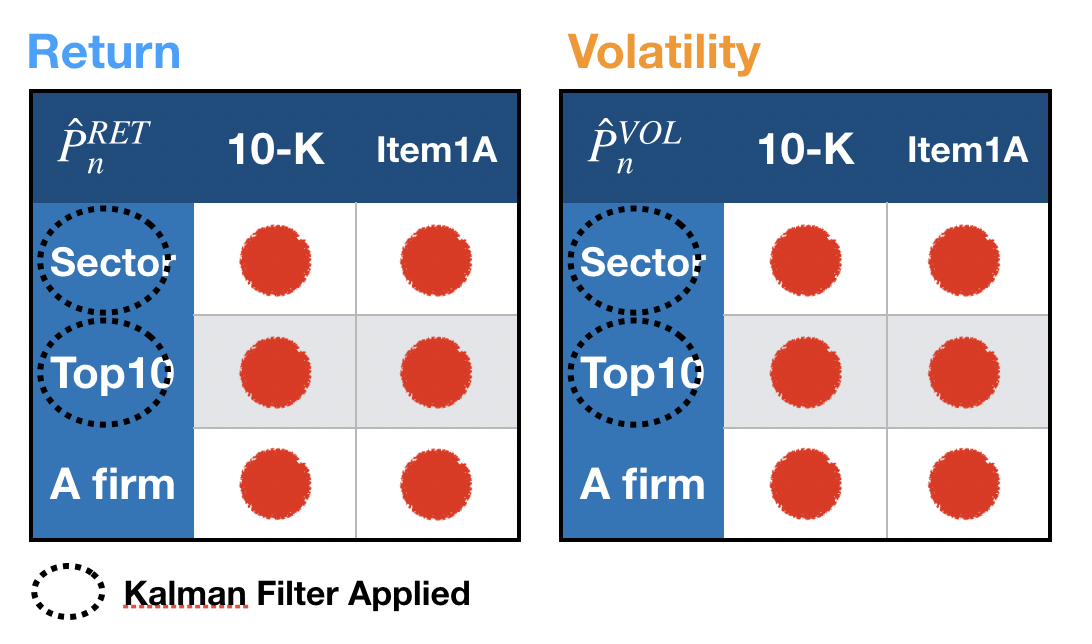}
\caption{The twelve sentiment metrics evaluated: three stakeholder levels
(sector, portfolio, firm) $\times$ two document variants (10-K, Item~1A)
$\times$ two supervisory targets (return, volatility). Kalman filtering is
applied at the sector and portfolio level only.}
\label{fig:twelvemetrics}
\end{figure}

For each of the three stakeholder levels -- sector (QQQ, all 94 firms),
portfolio (top-10 QQQ constituents), and firm (Nvidia) -- we train the
Sentiment Score Prediction Model separately on the full-10-K and Item~1A
document variants, against both return and volatility labels, yielding
twelve sentiment metrics in total ($2$ document variants $\times$ $2$
targets $\times$ $3$ levels; Figure~\ref{fig:twelvemetrics}). Sector- and
portfolio-level series are Kalman-filtered (Section~\ref{sec:methodology})
prior to analysis; firm-level series are analysed unfiltered.

We report three independent evaluations for each metric: (i)
\emph{classification accuracy and loss} of the sentiment score against the
return/volatility label it was trained on; (ii) \emph{Pearson correlation}
among the return-supervised score, the volatility-supervised score, a
Loughran--McDonald dictionary baseline $\tilde{p}^{LM}$, and the
corresponding price series; and (iii) a \emph{qualitative analysis} of the
top-15 words driving each model's positive and negative sentiment
classification.

\subsection{Full-10-K versus Item~1A: Accuracy and Loss}

\begin{table}[h]
\centering
\caption{Sentiment score classification accuracy and loss by level, document
variant, and target.}
\label{tab:ablation}
\begin{tabular}{llcccc}
\toprule
Level & Document & Acc.\ ($\hat{p}^{RET}$) & Loss & Acc.\ ($\hat{p}^{VOL}$) & Loss \\
\midrule
Sector              & Full 10-K & 78\% & 0.25 & 92\% & 0.22 \\
Sector              & Item 1A   & 73\% & 0.25 & 90\% & 0.22 \\
Portfolio (Top 10)  & Full 10-K & 76\% & 0.22 & 78\% & 0.20 \\
Portfolio (Top 10)  & Item 1A   & 71\% & 0.23 & 74\% & 0.21 \\
Firm (Nvidia)       & Full 10-K & 75\% & 0.19 & 69\% & 0.25 \\
Firm (Nvidia)       & Item 1A   & 76\% & 0.21 & 75\% & 0.25 \\
\bottomrule
\end{tabular}
\end{table}

The central ablation finding is that full 10-K text outperforms Item~1A
alone at both the sector and portfolio level, for both targets, by margins
of 2--5 percentage points. This reverses specifically at the individual-firm
level, where the Item~1A model outperforms the full-filing model by 1 point
on return and 6 points on volatility. We interpret this as an aggregation
effect: at the sector and portfolio level, the larger, more heterogeneous
vocabulary of the full filing gives the word-scoring step
(Eq.~\eqref{eq:fw}) more material from which to isolate genuinely
sentiment-charged terms across many firms' filings; at the single-firm
level, with only one firm's filings as training signal, the full filing's
boilerplate and financial-statement language dilutes the smaller set of
terms that carry real signal, whereas Item~1A's narrower vocabulary is a
better match to the amount of training signal available at $n=1$.

\begin{figure}[h]
\centering
\begin{subfigure}{0.48\textwidth}
\includegraphics[width=\textwidth]{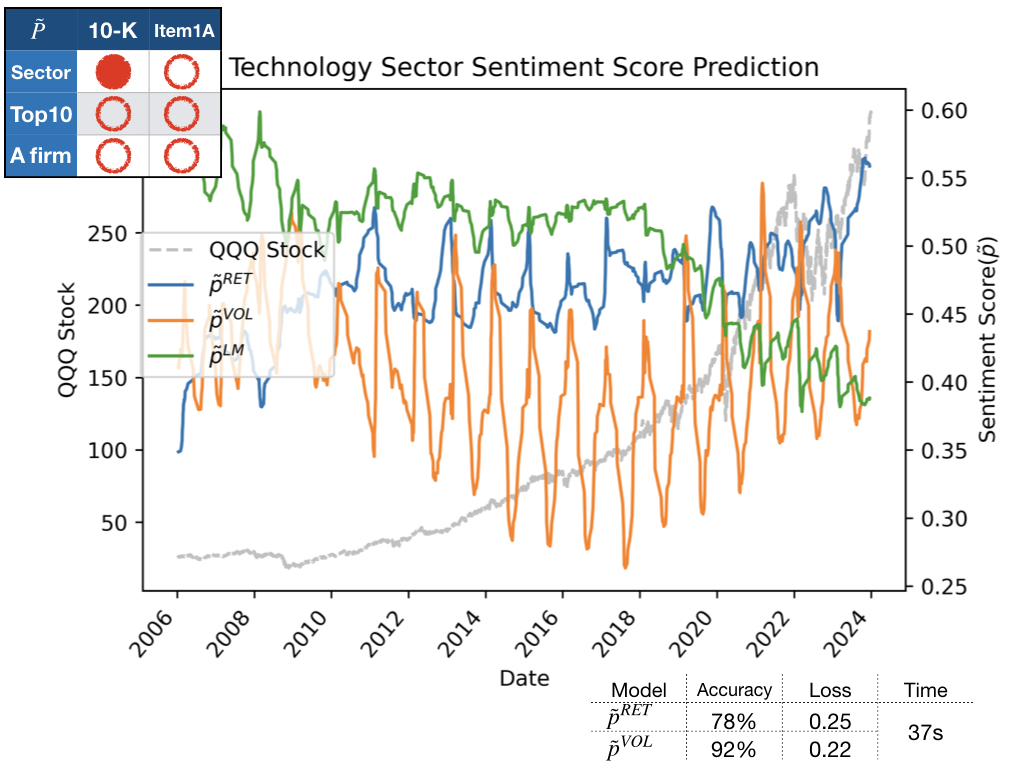}
\caption{Full 10-K}
\end{subfigure}
\hfill
\begin{subfigure}{0.48\textwidth}
\includegraphics[width=\textwidth]{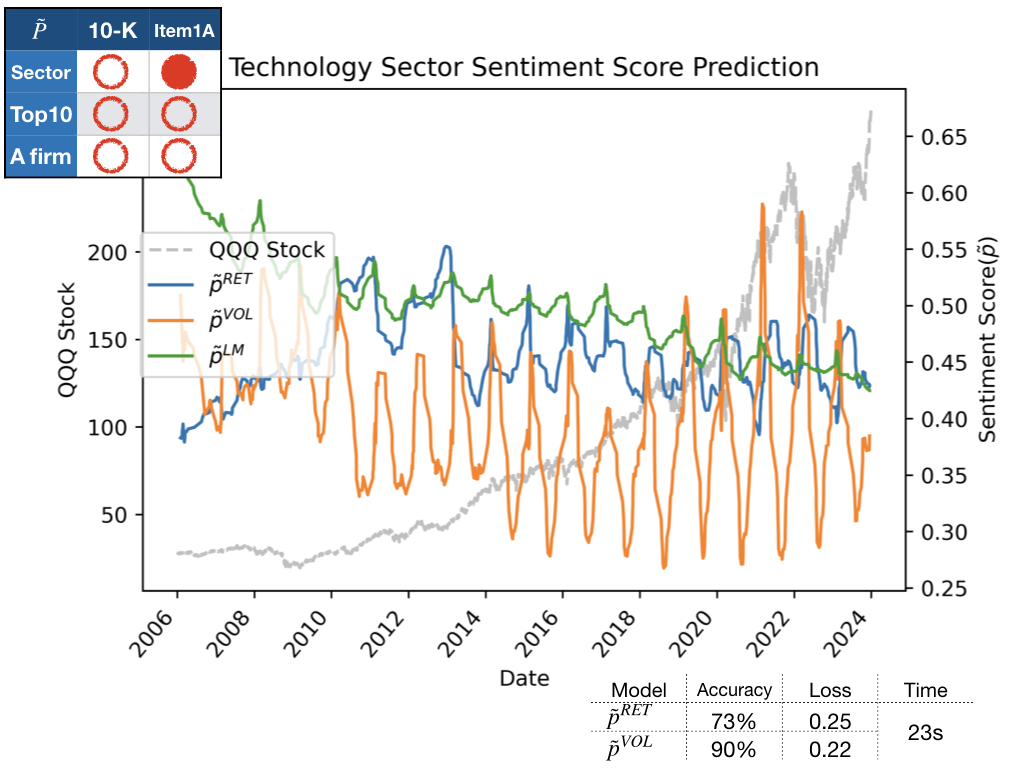}
\caption{Item 1A only}
\end{subfigure}
\caption{Sector-level (technology sector, QQQ) sentiment score prediction,
full 10-K versus Item~1A. $\tilde{p}^{RET}$, $\tilde{p}^{VOL}$: Kalman-filtered
return- and volatility-supervised scores; $\tilde{p}^{LM}$:
Loughran--McDonald baseline.}
\label{fig:sector_trends}
\end{figure}

\begin{figure}[h]
\centering
\begin{subfigure}{0.48\textwidth}
\includegraphics[width=\textwidth]{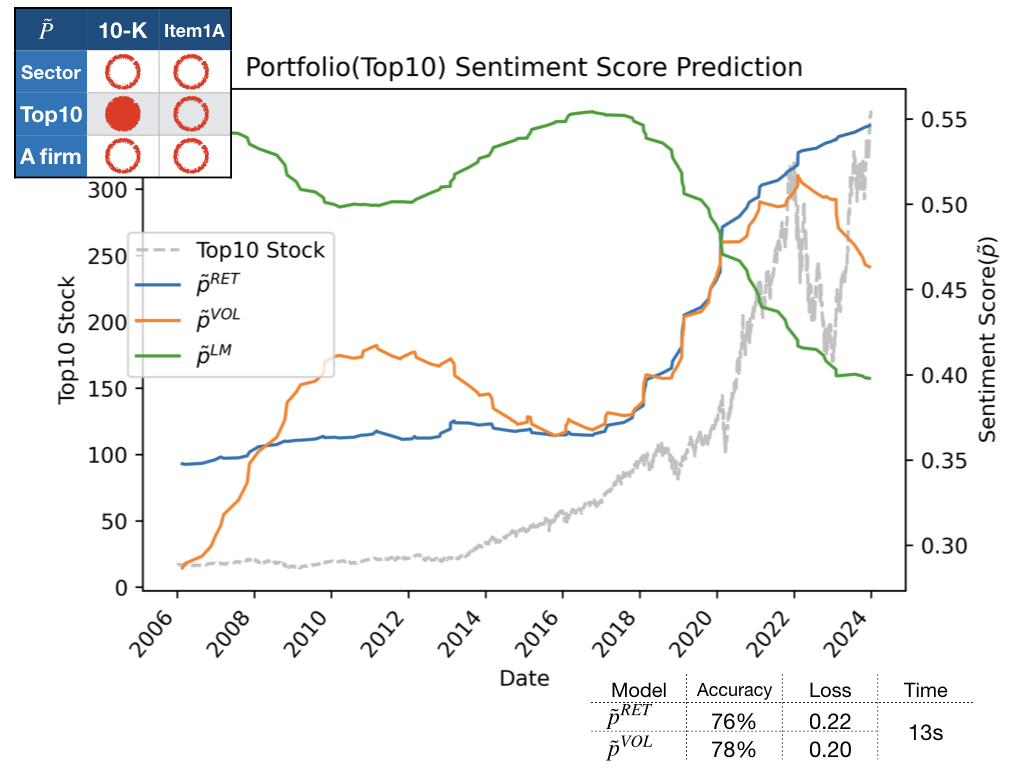}
\caption{Full 10-K}
\end{subfigure}
\hfill
\begin{subfigure}{0.48\textwidth}
\includegraphics[width=\textwidth]{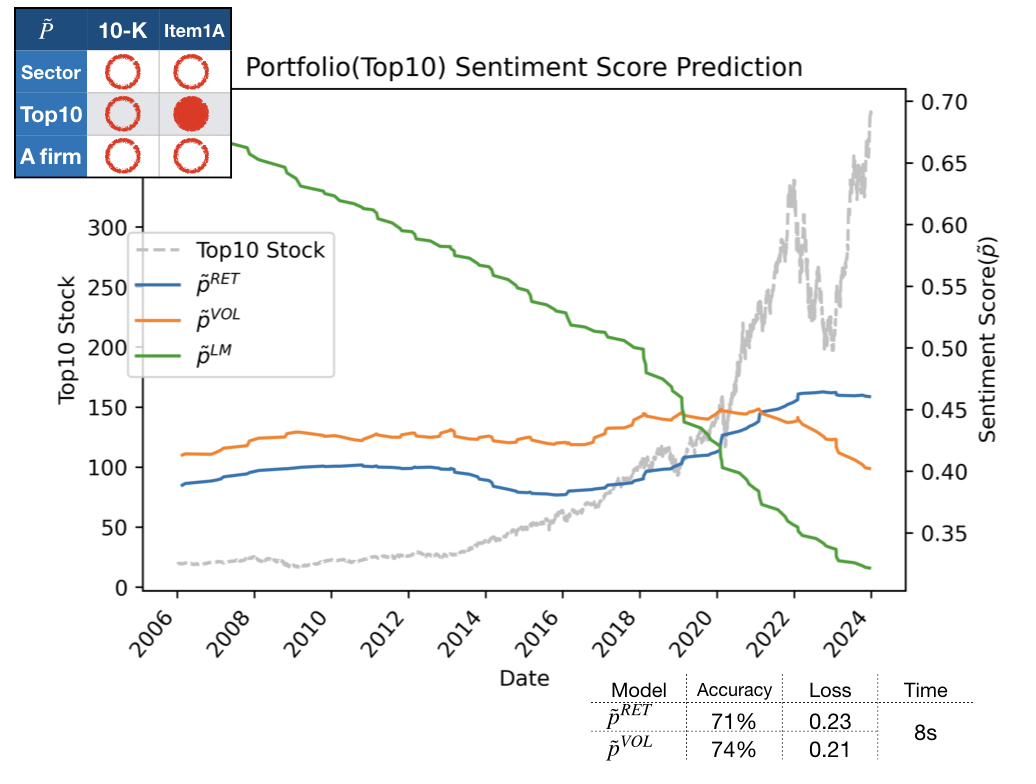}
\caption{Item 1A only}
\end{subfigure}
\caption{Portfolio-level (top-10 QQQ constituents) sentiment score
prediction, full 10-K versus Item~1A.}
\label{fig:portfolio_trends}
\end{figure}

\begin{figure}[h]
\centering
\begin{subfigure}{0.48\textwidth}
\includegraphics[width=\textwidth]{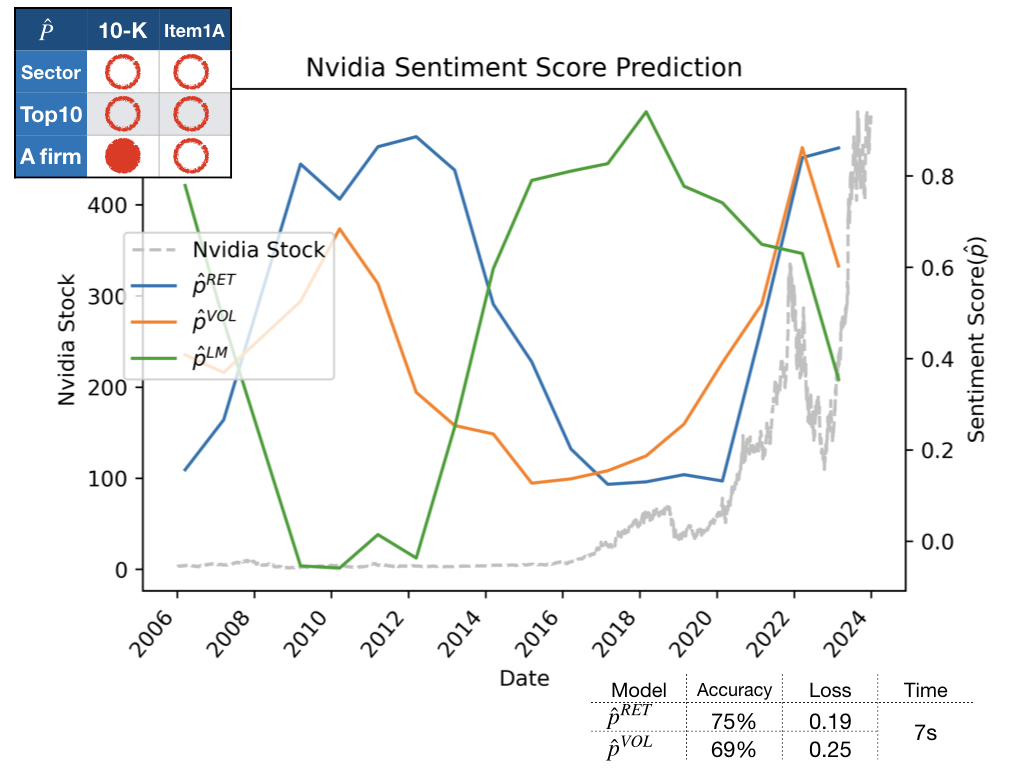}
\caption{Full 10-K}
\end{subfigure}
\hfill
\begin{subfigure}{0.48\textwidth}
\includegraphics[width=\textwidth]{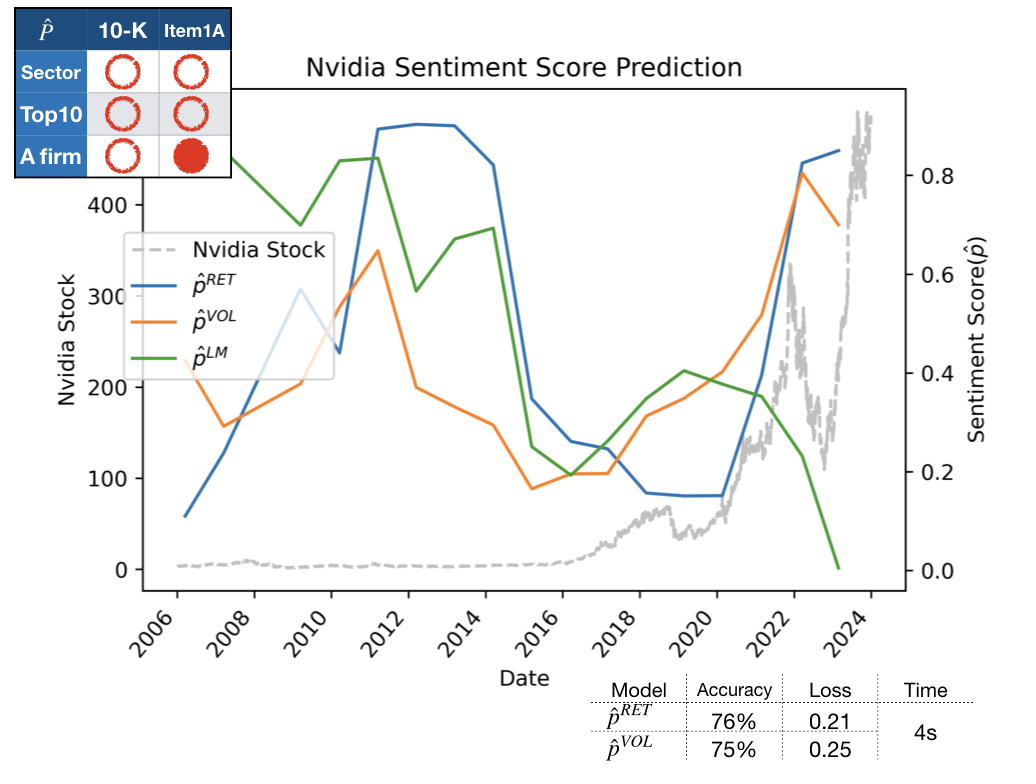}
\caption{Item 1A only}
\end{subfigure}
\caption{Firm-level (Nvidia) sentiment score prediction, full 10-K versus
Item~1A. Firm-level scores are unsmoothed (no Kalman filter).}
\label{fig:firm_trends}
\end{figure}

\subsection{Correlation Analysis}

\begin{figure}[h]
\centering
\includegraphics[width=\textwidth]{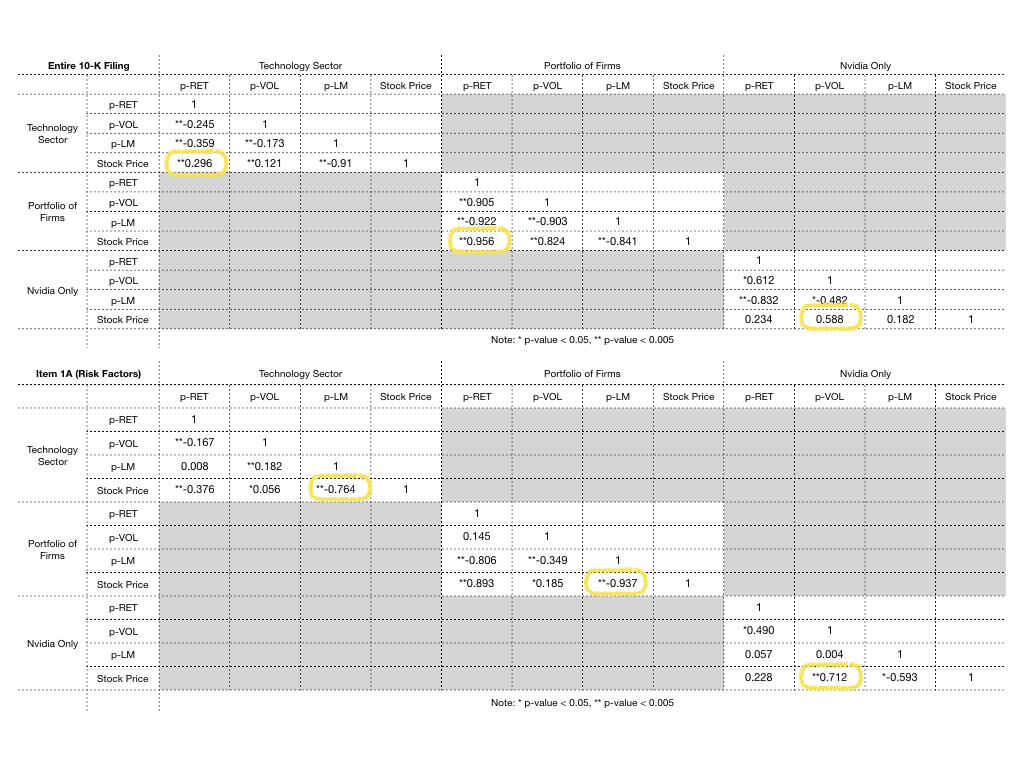}
\caption{Full Pearson correlation matrices across all six models (sector,
portfolio, and firm level; full 10-K and Item~1A document variants).
$p$-RET, $p$-VOL: return- and volatility-supervised sentiment scores;
$p$-LM: Loughran--McDonald baseline. $^{*}p<0.05$, $^{**}p<0.005$.}
\label{fig:correlation_table}
\end{figure}

Figure~\ref{fig:correlation_table} reports the full set of Pearson
correlations across all six models (three levels $\times$ two document
variants). At the sector level (full 10-K, filtered), return- and
volatility-supervised sentiment are negatively correlated ($r=-0.245$,
$p<0.005$); both are negatively correlated with the LM baseline
($r=-0.359$ and $-0.173$); the LM baseline itself is strongly negatively
correlated with QQQ price ($r=-0.91$), while return-supervised sentiment is
positively correlated with price ($r=0.296$).

At the portfolio level (Top-10), price correlates strongly and positively
with return sentiment under the full-10-K model ($r=0.956$) and moderately
under Item~1A ($r=0.893$), while the LM baseline again shows a strong
negative relationship with price under both variants ($r=-0.841$ and
$-0.937$ respectively) -- a pattern consistent across every level tested, in
which the dictionary-based baseline moves opposite to price.

At the firm level (Nvidia), the return- and volatility-supervised scores are
\emph{positively} correlated under both document variants -- $r=0.612$
(full 10-K) and $r=0.490$ (Item~1A), both significant at $p<0.05$ -- in
contrast to the negative RET--VOL relationship observed at the sector and
portfolio level. Price correlates positively with volatility sentiment
under both variants ($r=0.588$ full 10-K; $r=0.712$ Item~1A), while the LM
baseline's relationship with price flips sign between variants at the firm
level ($r=0.182$, not significant, under full 10-K; $r=-0.593$, significant,
under Item~1A) -- the only level at which the LM-price relationship is not
consistently negative, worth a dedicated remark in Discussion rather than
treating it as noise.

\subsection{Qualitative Word Analysis}

\paragraph{Sector, full 10-K.} Positive-return words cluster into two
themes: \emph{Innovation and Development} (\textit{musk}, \textit{maximizer},
\textit{searcher}, \textit{torque}, \textit{bolivia} -- the latter
interpreted via lithium's role in EV batteries \citep{chan2023global}) and
\emph{Financial Health} (\textit{gilt}, \textit{div}, \textit{memoranda},
\textit{stance}) \citep{koijen2016financial}. Negative-return words support a
\emph{Technological Difficulty} theme (\textit{transformer}, \textit{scalar},
\textit{tango}, \textit{tera} -- \textit{tango} referencing Google's
discontinued augmented-reality project
\citep{kastrenakes2017google,donfro2017google}). Volatility-increasing words
split into \emph{COVID-19} (\textit{lancet}, \textit{forearm}) and
\emph{Innovation and Growth} (\textit{fang}, \textit{granularity},
\textit{renter}).

\paragraph{Sector, Item~1A.} Negative-return words shift toward a distinct
\emph{Biotechnology} theme (\textit{biogen}, \textit{kinase},
\textit{chemotherapy}) plus an explicit \textit{covid} token. Volatility-increasing
words suggest \emph{Cybersecurity} (\textit{spoof}, \textit{impropriety})
and \emph{wearable healthcare technology} (\textit{clothing},
\textit{polymeric}).

\paragraph{Portfolio, full 10-K.} Both return- and volatility-increasing
word sets converge on \emph{Electric Cars} (\textit{battery},
\textit{solar}, \textit{roadster}, \textit{motor}; and
\textit{autonomous}, \textit{neural}, \textit{pilot}, \textit{ford}).
Volatility-decreasing words support \emph{Stability and Resilience}
(\textit{resilience}, \textit{tester}).

\paragraph{Portfolio, Item~1A.} The \emph{Electric Cars} theme persists for
return (\textit{roadster}, \textit{supercharger}, \textit{sedan}), but
Item~1A additionally surfaces a theme absent from the full-text model:
\emph{Supply Chain Risk} (\textit{subcontractor}, \textit{entrust},
\textit{dependence}, \textit{nationwide})
\citep{economist2020,tingfang2021,lewis2019,wu2021}. Volatility words split
into \emph{Innovation} (\textit{circuit}, \textit{virtual}, \textit{transact})
\citep{Mazzucato2012,Gharbi2014,Hai2020} and \emph{Infrastructure and
Services} (\textit{depot}, \textit{host}, \textit{azure},
\textit{subscriber}) \citep{Brinkman2022,BlancBrude2022,Gunnion2021}.

\paragraph{Firm (Nvidia), full 10-K.} Positive-return words support
\emph{Innovation and Development} (\textit{thermal}, \textit{talent},
\textit{connect}) and a firm-specific \emph{GPU Market Domination} theme
(\textit{chain}, \textit{tender}, \textit{exclusion})
\citep{leswing2023nvidia,nvidia2023supplychain}. Negative-return words
cluster around \emph{Market Challenges} (\textit{grid}, \textit{petition},
\textit{eastern}, \textit{enactment}).

\emph{Note on interpretation.} Word-to-theme mapping was performed by the
authors with reference to contemporaneous industry context; each proposed
theme is additionally corroborated by citation to prior literature
establishing the relevant relationship. The interpretation is suggestive
rather than a validated topic model, but each theme is anchored to an
independent, citable empirical claim rather than free association.

%% file: sections/discussion.tex
\section{Discussion}

\paragraph{The central finding, stated precisely.}
Full-10-K text produces more accurate, better-calibrated sentiment at the
sector and portfolio level, for both return and volatility targets, by
margins of 2--5 percentage points. This reverses at the individual-firm
level, where the Item~1A risk-factor section outperforms the full filing by
1 point on return and 6 points on volatility. We interpret this as an
aggregation effect rather than a property of the text itself: at the sector
and portfolio level, the model draws on 94 and 10 firms' filings
respectively, and the larger, more heterogeneous vocabulary of the full
filing gives the word-scoring step (Eq.~\eqref{eq:fw}) more material from
which to isolate genuinely sentiment-charged terms. At the single-firm
level, with only one firm's filings as training signal, the full filing's
boilerplate and financial-statement language -- sections unlikely to vary
meaningfully in tone from year to year for a single firm -- dilutes the
smaller set of terms that do carry real signal, whereas the risk-factor
section's narrower, more concentrated vocabulary is a better match to the
amount of training signal available at $n=1$.

\paragraph{A structural bias worth reporting plainly, not softening.}
The Loughran--McDonald dictionary baseline is negatively correlated with
price at every level we test (sector: $r=-0.910$; portfolio: $r=-0.841$ to
$-0.937$ across both document variants), a far stronger and more consistent
relationship than either supervised score shows in either direction. This is
not evidence the dictionary approach is ``wrong'' so much as evidence of a
known structural property: the LM dictionary is deliberately conservative
and skews negative by design, and 10-K risk-factor language is itself
disclosure-mandated and cautious by convention -- three-fourths of Harvard
IV-4-flagged negative words don't carry negative connotation in financial
text, which is exactly why \citet{LoughranMcDonald2011} built a
domain-specific list in the first place, but the domain-specific list still
leans negative on this corpus. We report this as a limitation of
dictionary-based scoring on regulatory text specifically, which strengthens
the case for the supervised approach used here.

\paragraph{Risk-factor text surfaces themes full-text does not.}
At both the portfolio and sector level, the Item~1A-only model recovers
themes absent from the full-filing model's word set -- \emph{Supply Chain
Risk} at portfolio level, an explicit \textit{covid} token and
\emph{Biotechnology} cluster at sector level. This suggests the risk-factor
section's value isn't purely in its concentration (fewer, more relevant
words) but in surfacing forward-looking risk narratives that get diluted by
volume in the full filing -- a point that complements, rather than just
competes with, the accuracy-based ablation result above.

\paragraph{Limitations.}
The model operates on a bag-of-words representation and cannot capture
multi-word phrases (e.g., ``chief executive officer'' is scored as three
independent unigrams, losing its compositional meaning) -- a natural
extension is bigram or $n$-gram support. Sector- and portfolio-level
aggregates also do not weight firms by their actual index allocation; an
equally-weighted aggregate treats a 9\%-weighted constituent identically to
a firm with under 1\% weight, which could materially shift the aggregate
sentiment trend if reweighted. Finally, thematic word interpretation, while
reference-supported (above), remains a human-authored reading of model
output rather than an independently validated topic labeling.

%% file: sections/conclusion.tex
\section{Conclusion}

We presented a supervised, lexicon-learning approach to sentiment generation
from 10-K filings, extending prior return-only, news-based methodology to
(i) regulatory filing text, (ii) volatility as a second supervisory target,
and (iii) three levels of stakeholder aggregation. Across sector, portfolio,
and firm-level analysis, full-filing text produces more accurate and
better-calibrated sentiment scores in aggregate, while the Item~1A
risk-factor section becomes the stronger choice specifically at the
individual-firm level -- a reversal we attribute to the interaction between
text volume and the amount of independent training signal available at each
level of aggregation, rather than to any inherent superiority of one
document variant. The Loughran--McDonald dictionary baseline shows a
strong, consistent negative correlation with price at every level tested,
underscoring the value of a supervised approach for regulatory disclosure
text specifically. These sentiment metrics, and the design choices they
motivate, form the basis for a subsequent, larger-scale, multi-source
extension of this methodology.

%% file: references.bib
@article{ke2020predicting,
  author={Ke, Zheng and Kelly, Bryan T. and Xiu, Dacheng},
  title={Predicting Returns with Text Data},
  journal={University of Chicago, Becker Friedman Institute for Economics Working Paper},
  number={2019-69},
  year={2020},
  month={September},
  note={Yale ICF Working Paper No. 2019-10, Chicago Booth Research Paper No. 20-37},
  url={https://ssrn.com/abstract=3389884},
  doi={10.2139/ssrn.3389884}
}

@article{LoughranMcDonald2011,
  title={When is a liability not a liability? Textual analysis, dictionaries, and 10-Ks},
  author={Loughran, Tim and McDonald, Bill},
  journal={The Journal of Finance},
  volume={66},
  number={1},
  pages={35--65},
  year={2011},
  publisher={Wiley Online Library}
}

@article{corsi2009simple,
  title={A simple approximate long-memory model of realized volatility},
  author={Corsi, Fulvio},
  journal={Journal of Financial Econometrics},
  volume={7},
  number={2},
  pages={174--196},
  year={2009},
  publisher={Oxford University Press}
}

@book{durbin2012time,
  title={Time series analysis by state space methods},
  author={Durbin, James and Koopman, Siem Jan},
  volume={38},
  year={2012},
  publisher={OUP Oxford}
}

@techreport{Blomme2020,
  author={Blomme, Sander and Dedeyne, Julie},
  title={Predicting the effect of 10-K, 10-Q and 8-K company reports on abnormal stock returns using FinBERT NLP methods},
  year={2020},
  institution={University of Ghent},
  url={https://libstore.ugent.be/fulltxt/RUG01/002/837/812/RUG01-002837812_2020_0001_AC.pdf}
}

@misc{sec2005,
  title={Securities and Exchange Commission Final Rule, Release No. 33--8591 (FR-75)},
  author={{Securities and Exchange Commission}},
  year={2005},
  howpublished={\url{http://sec.gov/rules/final/33-8591.pdf}}
}

@article{economist2020,
  title={America's War on Huawei Nears Its Endgame},
  author={{The Economist}},
  year={2020},
  month={July},
  day={16},
  journal={The Economist},
  url={https://www.economist.com/briefing/2020/07/16/americas-war-on-huawei-nears-its-endgame}
}

@article{tingfang2021,
  title={US-China Tech War: Beijing's Secret Chipmaking Champions},
  author={Ting-Fang, C. and Li, L.},
  year={2021},
  month={May},
  day={5},
  journal={Nikkei Asia},
  url={https://asia.nikkei.com/Spotlight/The-Big-Story/US-China-tech-war-Beijing-s-secret-chipmaking-champions}
}

@book{lewis2019,
  title={Learning the superior techniques of the barbarians: China's pursuit of semiconductor independence},
  author={Lewis, J.},
  year={2019},
  publisher={CSIS},
  address={Washington}
}

@article{wu2021,
  title={Chip Shortage Set to Worsen as Covid Rampages Through Malaysia},
  author={Wu, D. and Lee, Y. and Ngui, Y.},
  year={2021},
  month={August},
  day={23},
  journal={Bloomberg},
  url={https://www.bloomberg.com/news/articles/2021-08-23/chip-shortage-set-to-worsen-as-covid-rampages-through-malaysia}
}

@article{Mazzucato2012,
  title={R\&D, patents and stock return volatility},
  author={Mazzucato, M. and Tancioni, M.},
  journal={Journal of Evolutionary Economics},
  volume={22},
  pages={811--832},
  year={2012},
  doi={10.1007/s00191-012-0289-x}
}

@article{Gharbi2014,
  title={R\&D investments and high-tech firms' stock return volatility},
  author={Gharbi, Sami and Sahut, Jean-Michel and Teulon, Fr{\'e}d{\'e}ric},
  journal={Technological Forecasting and Social Change},
  volume={88},
  pages={306--312},
  year={2014},
  doi={10.1016/j.techfore.2013.10.006}
}

@article{Hai2020,
  title={R\&D volatility and market value: the role of executive overconfidence},
  author={Hai, B. and Gao, Q. and Yin, X. and Chen, J.},
  journal={Chinese Management Studies},
  volume={14},
  number={2},
  pages={411--431},
  year={2020},
  doi={10.1108/CMS-05-2019-0170}
}

@misc{Brinkman2022,
  title={Infrastructure investing will never be the same},
  author={Brinkman, Marcel and Sarma, Vijay},
  year={2022},
  month={8},
  howpublished={McKinsey \& Company}
}

@misc{BlancBrude2022,
  title={Infrastructure Strategy 2022: A Pivot to the Digital Frontier},
  author={Blanc-Brude, Fr{\'e}d{\'e}ric and Schmundt, Wilhelm and Bumberger, Thomas and Friedrich, Roman and Georgii, Bernhard and Gupta, Abhishek and Lum, Leonard and Wilms, Maikel},
  year={2022},
  month={3},
  howpublished={Boston Consulting Group}
}

@misc{Gunnion2021,
  title={Infrastructure investment: An economist's view from the ground up},
  author={Gunnion, Lester},
  year={2021},
  month={7},
  journal={Economics Spotlight},
  publisher={Deloitte}
}

@misc{chan2023global,
  author={Chan, Luke and Devia-Valbuena, Nicol{\'a}s},
  title={In the Global Rush for Lithium, Bolivia is at a Crossroads},
  year={2023},
  month={12},
  day={12},
  url={https://www.usip.org/publications/2023/12/global-rush-lithium-bolivia-crossroads}
}

@article{koijen2016financial,
  title={Financial Health Economics},
  author={Koijen, Ralph S.J. and Philipson, Tomas J. and Uhlig, Harald},
  journal={Econometrica},
  volume={84},
  pages={195--242},
  year={2016},
  doi={10.3982/ECTA11182}
}

@misc{kastrenakes2017google,
  author={Kastrenakes, Jacob},
  title={Google's Project Tango is shutting down because ARCore is already here},
  year={2017},
  month={12},
  day={15},
  url={https://www.theverge.com/2017/12/15/16782556/project-tango-google-shutting-down-arcore-augmented-reality}
}

@misc{donfro2017google,
  author={D'Onfro, Jillian},
  title={Google is going to shut down its once-hyped augmented reality project, Tango},
  year={2017},
  month={12},
  day={15},
  journal={CNBC},
  url={https://www.cnbc.com/2017/12/15/google-kills-project-tango-ar-project.html}
}

@misc{leswing2023nvidia,
  title={Meet the \$10,000 Nvidia chip powering the race for A.I.},
  author={Leswing, Kif},
  year={2023},
  howpublished={\url{https://www.cnbc.com/2023/02/23/nvidias-a100-is-the-10000-chip-powering-the-race-for-ai-.html}}
}

@misc{nvidia2023supplychain,
  author={{NVIDIA}},
  title={NVIDIA Industries: Retail - Supply Chain Management},
  year={2024},
  howpublished={\url{https://www.nvidia.com/en-gb/industries/retail/supply-chain-management/}}
}
